# Emotions, Context, and Substance Use in Adolescents: A Large Language Model Analysis of Reddit Posts

Jianfeng Zhu[1], Hailong Jiang[1], Yulan Wang[3], Karin G. Coifman[2], Ruoming Jin [1], Deric R. Kenne[4]

1 Department of Computer Science, Kent State University, Kent, OH 44224
2 Department of Psychological Sciences, Kent State University, Kent, OH 44224
3 Northeastern University Khoury College of Computer Sciences, Boston, MA 02115
4 College of Public Health, Kent State University, Kent, OH 44224

*Jianfeng Zhu.
**Email:** jzhu10@kent.edu

## Abstract

**Background:** Early substance use in adolescence increases risk for later substance use disorders and mental health problems, yet the real-time emotional and contextual factors underlying these behaviors remain poorly understood.
**Methods:** We analyzed 23,000 substance use–related posts and an equal number of non–substance use posts from Reddit's r/teenagers (2018–2022). Posts were annotated for discrete emotions (sadness, anger, joy, guilt, fear, disgust) and contextual influences (family, peers, school) using large language models (LLMs). Statistical tests compared group differences, heatmaps visualized emotion–context co-occurrences, and interpretable machine learning models (SHAP) identified key predictors of substance use discussions. In addition, LLM-based thematic analysis extracted latent psychosocial themes linking emotions with contexts, providing qualitative insight into adolescents lived experiences.
**Results:** Negative emotions, particularly sadness, guilt, fear, and disgust, were more prevalent in substance use posts, while joy predominated in non–substance use discussions. Guilt and shame diverged: guilt often signaled regret and reparative intent, whereas shame reinforced risky dynamics through peer identity performance. Peer influence was the strongest contextual driver, closely tied to sadness, fear, and guilt. In contrast, family and school environments displayed dual roles, acting as both risk and protective contexts depending on relational quality and stressors.
**Conclusions:** Adolescent substance use reflects the interplay of sadness, guilt, and peer influence as central risk factors, with family and school contexts offering both risk and protection. By combining statistical tests, interpretable machine learning, and LLM-assisted thematic coding, this study shows the utility of mixed-computational approaches in clarifying the nuanced mechanisms linking emotion, context, and adolescent risk.

## Introduction

Adolescent substance use (SU) is a pressing public health concern. in 2023, at least one in eight teenagers reported using an illicit substance in the past year, SU among 8th graders rose by 61% between 2016 and 2020, and more than 62% of 12th graders reported alcohol misuse [1]. Adolescence is a critical developmental window in which initial experimentation on occurs, with tobacco and alcohol often preceding illicit drug use [2]. A range of factors contribute to substance use in teenagers, including peer influence, family dynamics, school performance, and emotional states shape adolescents' risk profile [3,4]. While these risk factors are well recognized, the real-time interplay between emotions and contextual experiences surrounding SU remains poorly understood. Deepening our understanding of these moment-to-moment relationships is essential for informing educational strategies and designing timely, targeted interventions that resonate with adolescents' daily realities.

Conventional surveys and longitudinal assessments provide valuable population estimates but are time-lagged, burdensome, and coarse with respect to rapidly shifting feelings and social contexts. Emotions states can fluctuate within minutes or hours, and exposure to peers, school stressors, or family conflict are often episodic; these dynamics are difficult to capture with periodic questionnaires or clinic-based assessments. Ambulatory assessment methods, such as ecological momentary assessment and passive digital data collection, have been used to capture "real-world" microprocesses in youth [5] offering granular insight into affective and behavioral fluctuations in daily life.

Social media platforms often a complementary approach, capturing the dynamic, bidirectional interplay between adolescents' behavior and contextual events. Adolescents use these platforms to share immediate mood states and narrate experiences embedded within peer, school, and family settings [6–8]. Prior studies have leverages social media data to track opioid-related content and other SU topics at scale, enabling near–real-time surveillance of trends and risk factors [9,10]. Such passive or self-reported data provide a unique opportunity to observe adolescent thoughts, feelings, and behaviors as they occur in real time, in naturalistic settings [11].

Recent advances in artificial intelligence, particularly large language models (LLMs) such as GPT-4 and Gemini, enable scalable, domain-adaptable annotation of text, including emotion detection and extraction of contextual themes [12,13]. Building on these advances, this study analyzed adolescent discussions on Reddit to: (i) apply LLMs to annotate each post with the specific emotions and contextual factors (family, peer, and school); (ii) quantify emotion context co-occurrence patterns via heatmaps; and (iii) identify predictors of SU-related posts using interpretable machine learning; and (iv) extract latent themes linking emotions with contexts through LLM-based thematic analysis. By connecting dynamic emotions with situated social contexts, this study aimed to detect actionable signals to inform the development of prevention strategies and timely interventions for adolescent SU.

**Literature review**

### Social Media Big Data for Substance Use Research

Social media big data refers to large scale, digitized resources comprising occurring communication on platforms such as Reddit, Twitter, and Facebook. These platforms enable researchers to observe real world phenomena at scale, providing rich, data-driven insights through computational analyses. Individuals often share personal experiences, questions, and reflections on substance use online, while those struggling with addiction often seek social support through these communities [14–16]. By analyzing user-generated content, researchers can extract valuable to both business and academic applications [17].

Compared with traditional surveys, social media offers scalable, real-time access to youth perspectives, including those from hard-to-reach groups such as homeless adolescents. This approach supports identification of substance use patterns and informs prevention and intervention strategies [27]. Sentiment analysis has been used to assess public attitudes toward substances, such as the varied emotional responses to synthetic opioids like fentanyl [19]. Moreover, the predominance of positive portrayals of substance use online can influence adolescent attitudes, underscoring the need for balanced prevention messaging [20].

### Emotional and Contextual Factors in Adolescent Substance Use

The relationship between emotions (e.g., sadness, shame, anger, joy, guilt) and adolescent substance use is complex. Sadness and shame, in particular, have been shown to strongly predict substance involvement [21]. Emotional distress is recognized as a mechanism in both the onset and maintenance of substance use disorders [22], while interventions that target emotion regulation and resilience can mitigate reliance on substances as coping tools [23,24].

Contextual influences also play a central role. Peer behaviors and perceived peer norms strongly shape adolescent choices, with pressure and fear of exclusion prompting experimentation [25]. Peer pressure, along with the fear of social exclusion or rejection, can exacerbate emotional and behavioral health issues, prompting adolescents to use substances as a coping mechanism or to seek acceptance in alternative peer groups [26,27]. Family dynamics—such as neglectful parenting and parental alcohol use—are linked to higher rates of initiation [15,18]. School environments, including teacher relationships, school satisfaction, and academic achievement, further influence substance use trajectories, either as risks or protective factors [29,30].

### LLMs and Machine Learning Approaches

Advances in natural language processing and large language models (LLMs) have opened new avenues for analyzing substance use patterns in digital data. Studies show that zero-shot and few-shot models can achieve accuracy comparable to human annotators in tasks such as stance detection, open-text coding, and survey labeling [8,31–33]. Model performance, however, is sensitive to prompt design, including length, output format, and definition framing [34].

Beyond LLMs, machine learning methods have been applied to mental health research for prediction and feature extraction. For instance, regression models with SHAP visualization have predicted adolescent anxiety onset by identifying key predictors [35]. Multi-view learning approaches that integrate diverse social media interactions (e.g., Facebook likes, status updates) have also achieved high accuracy in predicting substance use [36].

However, prior studies on adolescent substance use have typically examined emotions, family, peer, or school influences in isolation, leaving gaps in understanding their interplay in real-world digital contexts. Traditional surveys further miss the immediacy of youth expression, while few studies have leveraged large language models to annotate emotional and contextual factors at scale. Moreover, existing machine learning models often overlook these nuanced signals, limiting both interpretability and the design of targeted interventions.

## Research Questions

Building on evidence that emotions (e.g., sadness, shame), family dynamics, peer influence, and school environments are central to adolescent substance use, this study examines how these factors manifest in digital social spaces, specifically Reddit's r/teenagers. Prior research has often treated these dimensions separately, leaving gaps in understanding their interplay, especially through real-time youth expression. Moreover, few studies have applied LLMs to annotate emotional and contextual factors in adolescent populations.

Accordingly, this study addresses the following research questions:

**RQ1:** What emotions are most prevalent in adolescent substance-use posts, and how do they differ from non–substance-use posts?

**RQ2:** How do family, peer, and school contexts appear in substance-use posts, and how do these patterns differ from non–substance-use posts?

**RQ3:** How are emotions and contextual factors interrelated within substance-use and non–substance-use discussions?

**RQ4:** To what extent do emotional and contextual features predict substance-use content in machine learning models?

**RQ5.** How can large language models (LLMs) extract latent subthemes from emotion–context pairings, and what additional insights do these provide beyond quantitative analyses?

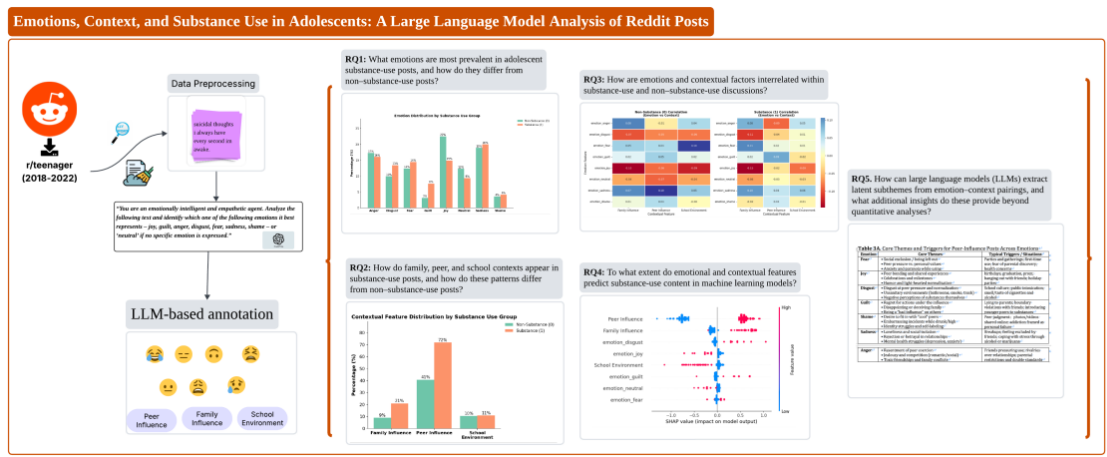

**Figure 1.** Study overview and representative findings

At a high level, our analytic pipeline is illustrated in Figure 1. The study is guided by five research questions (RQ1–RQ5), spanning emotion prevalence, contextual influences, and LLM-powered thematic analysis. By integrating Reddit data, LLM-based annotation, and machine learning, we aim to clarify the interplay between emotion and context in adolescent substance use.

## Methods

### Data Collection and Preprocessing

We analyzed user-generated posts from Reddit's r/teenagers (2018–2022), a forum with over 3.3 million members. Substance-use–related posts were identified using a curated set of keywords spanning multiple substances (e.g., alcohol, nicotine, cannabis, prescription drugs) [37]. For each post, the text fields (*title* and *selftext*) were concatenated into a single string. Posts with fewer than 10 words after preprocessing were excluded, as very short posts (e.g., "Yes," "That's cool") typically lack meaningful content. All text was lowercased, and stop words and punctuation were removed.

To verify the accuracy of our keyword filtering, we manually reviewed a random sample of 300 posts, achieving >90% precision in capturing substance-use content. Initially, we selected 23,275 substance-use posts and an equal number of non–substance-use posts sampled from the same time period to balance the dataset. After further data cleaning and preprocessing, the final analytic sample comprised 21,169 substance-use posts and 17,781 non–substance-use posts, for a total of 38,950 posts. Table 1 presents representative examples of substance-use posts alongside the matched keywords.

*Table 1* presents representative examples of substance-use posts alongside the matched keywords.

| Filtered Posts | Matched Term |
|---|---|
| i'm allowed to take 1-2mg of xanax when i get a severe panic attack. but even the xanax doesn't work that well. | xanax |
| suicidal thoughts i always have every second im awake. no amount of weed or alcohol or nicotine or other drugs help. | weed, alcohol, nicotine |
| bro some people got in a fight at school, so i went home early i also had like so many more cigarettes than i usually do in an entire month, holy fucking shit | cigarettes |

### LLM-based annotation of emotions and contexts factors

We utilized GPT-4 [38] as a prompt-based annotator to assign one primary emotion and up to three contextual labels for each post. This approach allows scalable yet nuanced content analysis, and prior studies show LLM annotations align well with human coders [39–41].

**Emotions (single label).** We adopted seven discrete emotions defined by the International Survey on Emotion Antecedents and Reactions: *joy, guilt, anger, disgust, fear, sadness,* and *shame* [42]. Posts without a clear emotion were labeled *neutral*. Example prompt: *"You are an emotionally intelligent and empathetic agent. Analyze the following text and*

*identify which one of the following emotions it best represents – joy, guilt, anger, disgust, fear, sadness, shame – or 'neutral' if no specific emotion is expressed."*

**Contexts (multi-label).** Each post was independently coded for the presence of *family*, *peer*, and *school* influences. Separate prompts were designed for each [43]. For example, the peer prompt asked: *"Determine whether the post shows any influence of peers (e.g., peer pressure, wanting to fit in, friends' behaviors). If yes, output 'Peer Influence' and briefly describe the peer-related context; if not, output 'None.'"*

### Statistical & Machine Learning Analyses

We conducted chi-square tests and independent-samples t-tests to compare the distributions and mean endorsement rates of emotional and contextual features between substance-use and non–substance-use posts (see Tables A and B). To visualize associations between emotions and contextual factors within substance-use posts, we generated a heatmap of co-occurrences.

To assess predictive power, we trained a binary XGBoost classifier (gradient-boosted decision trees) [44] to distinguish substance-use posts from non-substance-use controls, using the emotion and context labels as input features. Model performance was evaluated on a held-out test set (30% of the data) after a random train–test split, using accuracy, precision, recall (sensitivity), and F1 score as evaluation metrics.

To interpret model predictions, we examined both feature importance scores from XGBoost and Shapley Additive Explanations (SHAP) [45], which quantify each feature's marginal contribution to the model's prediction, enabling identification of the most influential emotional and contextual factors for substance-use detection in text posts.

### LLM-Based Thematic Extraction of Emotion–Context Subtopics

Beyond quantitative patterns, we sought to understand why adolescents express certain emotions in specific contexts around substance use. We therefore conducted a thematic analysis using GPT-4 to inductively extract deeper subtopics for each salient emotion × context pairing. Recent work suggests that LLMs can efficiently summarize and interpret large text collections to reveal themes deemed reasonable by experts[46], though some care is needed to ensure consistency. Our procedure was as follows:

**Sampling:** For each emotion–context combination of interest (e.g., *sadness × peer influence*, *guilt × family*, etc.), we retrieved up to 80 posts that were relatively rich in content (at least 100 words) to provide sufficient context. This ensured we analyzed substantive discussions rather than trivial posts.

**Prompting:** We fed the posts in batches to GPT-4 with an instruction to identify recurring subthemes or storylines, focusing on (i) reasons behind the expressed emotion, (ii) patterns in the contextual domain, and (iii) any stated triggers for substance use. For example, for *sadness × peer influence*, the prompt was:

*"Here are several posts from teens expressing sadness in situations involving peer influence (e.g., friends or classmates) related to substance use. Summarize the common themes or issues they talk about. What are recurring reasons they feel sad? How do peers figure into their stories? Are there common triggers for substance use mentioned?"*

**Synthesis:** GPT-4 outputs were manually checked by the first author to ensure coherence and accuracy. Representative subthemes and typical triggers were retained for each emotion–context pairing.

### Ethical Considerations

This study used publicly available, de-identified Reddit data. All personal identifiers (e.g., usernames) were removed. Because the data were publicly accessible and no direct interaction with human subjects occurred, institutional review board (IRB) approval and informed consent were not required. Findings are reported in aggregate to protect user privacy.

### Results

#### Temporal Trends in Substance Use Discussion

**Figure 2** shows the distribution of substance use-related posts in the r/teenagers subreddit from 2018 to 2022. Peaks were detected using a statistical method that identifies months when the proportion of substance-related posts is noticeably higher than both the previous and following months. A threshold ("prominence") of 0.2 percentage points was applied to ensure that only substantial increases were marked as peaks, thus minimizing the influence of minor fluctuations and highlighting more meaningful trends.

The frequency of posts increased steadily from March 2018, reached a pronounced peak in July 2019, and then declined sharply in August 2019, followed by a gradual decrease with minor fluctuations through the end of 2022. A chi-square test indicated seasonal variation in posting frequency, $\chi^2(3) = 178.91$, $p < .0001$, with higher activity in summer and holiday periods (seasonal proportions in Appendix Figure A1).

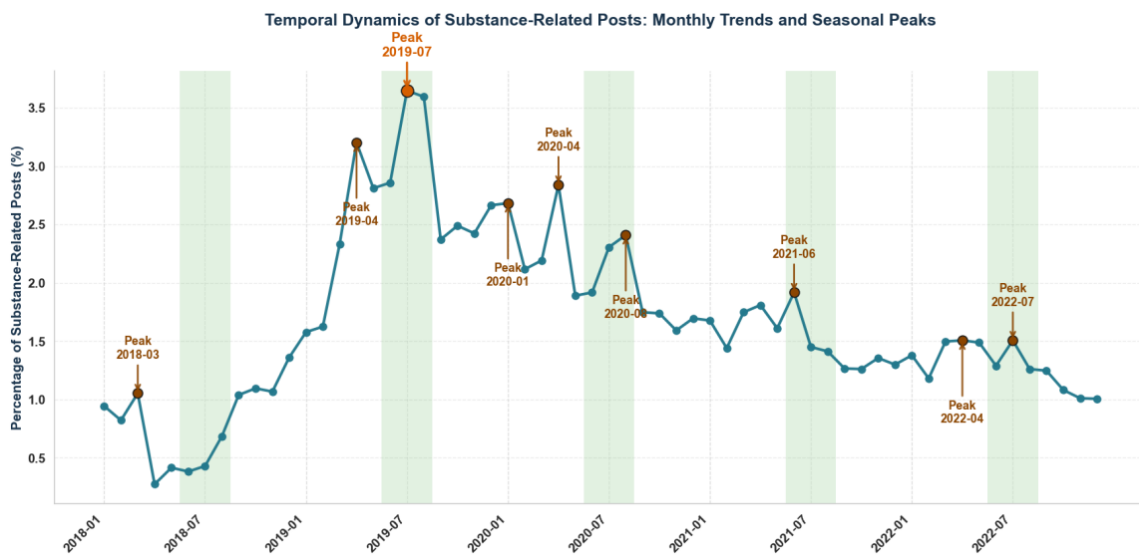

**Figure 2.** Monthly trends in the percentage of substance-related posts from 2018 to 2022.

## Emotion and Context Distributions by Substance Use Group

**Figure 3** displays the distribution of emotions in substance use versus non-substance use posts. Overall, sadness, joy, and anger were the most prevalent emotions across both groups. Notably, joy was substantially higher in non-substance use posts (22%) compared to substance use posts (15%), representing the largest relative difference. Sadness and anger were similarly common in both groups. In contrast, disgust, fear, and guilt were all more frequently expressed in substance use posts, with guilt showing a particularly pronounced difference (8% vs. 3%). Shame appeared with equal frequency (4%) in both groups.

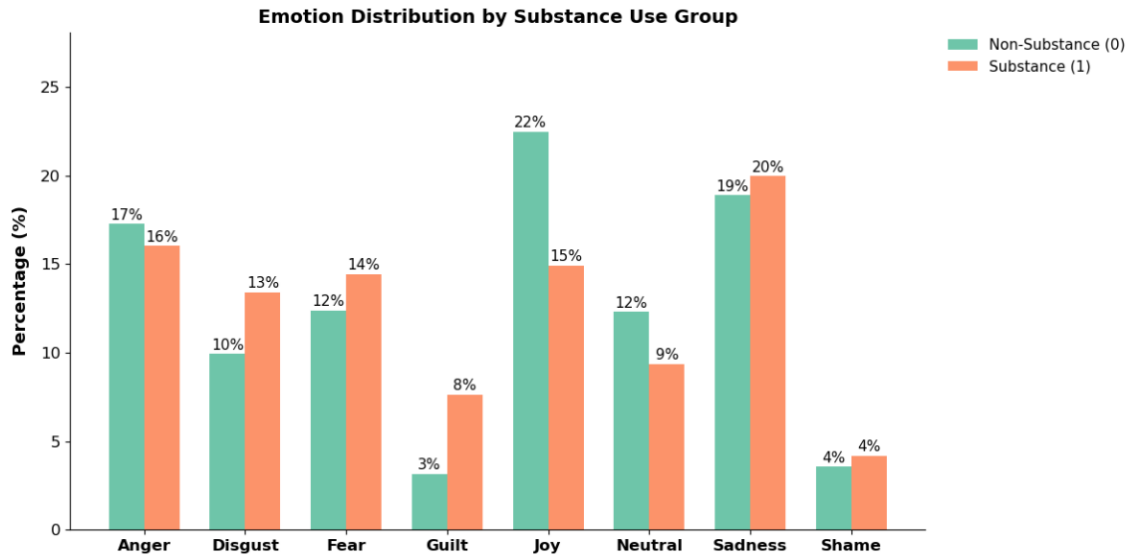

**Figure 3.** Emotion distribution in substance use versus non-substance use posts.

**Figure 4** displays the distribution of contextual influences across substance use and non-substance use posts. Peer influence was the most frequently mentioned context in both groups, with a notably higher proportion in substance use posts (72%) compared to non-substance use posts (41%). Family influence was also more common in substance use posts (21% vs. 9%). In contrast, school environment was discussed at a similar low frequency in both groups (around 10%).

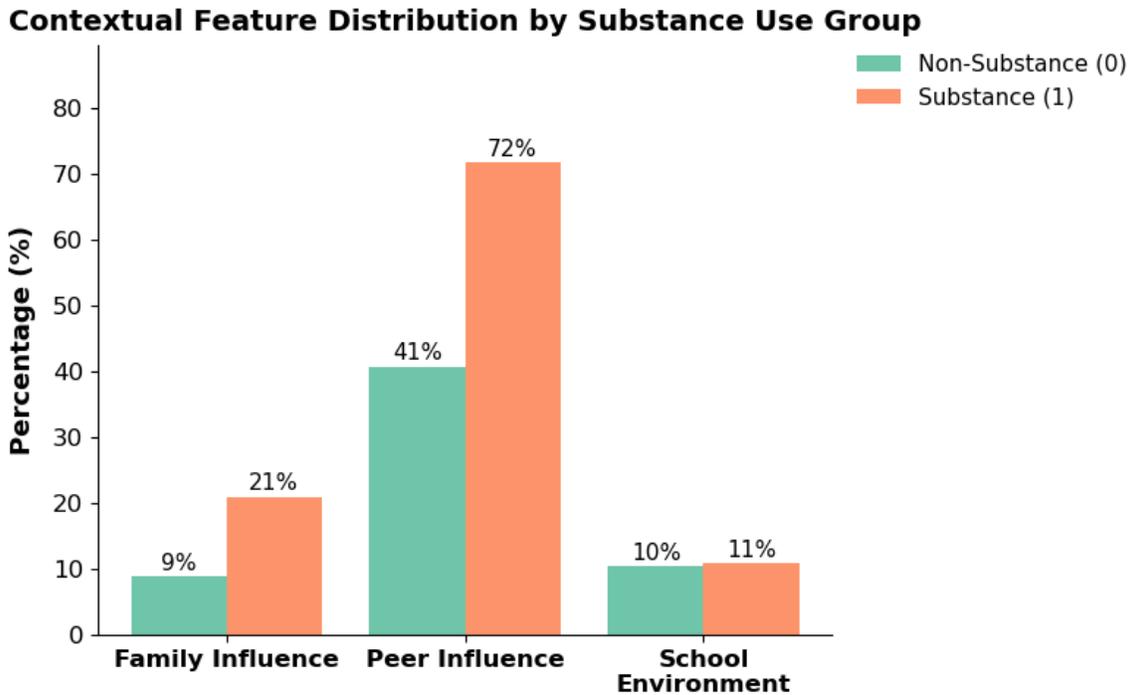

**Figure 4.** Contextual influences in substance use versus non-substance use posts.

### Statistical Comparisons and Associations

Group comparisons revealed significant differences in both emotional and contextual features between substance use and non-substance use posts. As shown in **Table 2A**, substance use posts contained higher levels of guilt, disgust, fear, sadness, and shame, whereas joy was significantly higher in non-substance use posts. Anger was slightly more common in non-substance use posts, though the effect size was small. Most emotional features showed highly significant group differences (p < .001), with anger (p = .001) and sadness (p = .008) also reaching significance at the p < .01 level.

As shown in **Table 2B**, substance use posts referenced peer influence and family influence significantly more often than non-substance use posts, with peer influence showing the largest difference between groups (p < .001). In contrast, references to the school environment did not differ significantly across groups (p = .185).

**Table 2A.** Emotional Feature Comparison by Substance Use Group

| Emotion | Substance Use (N=21169) M (SD) | Non-Substance Use (N=17781) M (SD) | t | p |
|---|---|---|---|---|
| **Joy** | 0.149 (0.356) | 0.225 (0.417) | -18.998 | <.001 |
| **Guilt** | 0.076 (0.266) | 0.032 (0.175) | 19.931 | <.001 |
| **Anger** | 0.160 (0.367) | 0.173 (0.378) | -3.260 | 0.001 |
| **Disgust** | 0.134 (0.341) | 0.099 (0.299) | 10.769 | <.001 |
| **Fear** | 0.145 (0.352) | 0.124 (0.329) | 5.963 | <.001 |
| **Sadness** | 0.200 (0.400) | 0.189 (0.392) | 2.671 | 0.008 |
| **Shame** | 0.042 (0.200) | 0.036 (0.185) | 3.221 | 0.001 |

**Table 2B.** Contextual Feature Comparison by Substance Use Group

| Contextual Feature | Substance Use (N=21169) M (SD) | Non-Substance Use (N=17781) M (SD) | t | p |
|---|---|---|---|---|
| Family Influence | 0.210 (0.407) | 0.088 (0.283) | 34.610 | <.001 |
| Peer Influence | 0.716 (0.451) | 0.407 (0.491) | 64.316 | <.001 |
| School Environment | 0.108 (0.311) | 0.104 (0.306) | 1.326 | 0.185 |

As show in **Figure 5,** emotion–context correlations differed across groups. In substance use posts, family influence was most strongly associated with fear and sadness, with stronger effects than in non-substance use posts. Peer influence showed the largest positive association with guilt, while school environment was most strongly linked to sadness. By contrast, in non-substance use posts, the school environment was most strongly correlated with fear. For both groups, family influence exhibited a consistent positive association with anger.

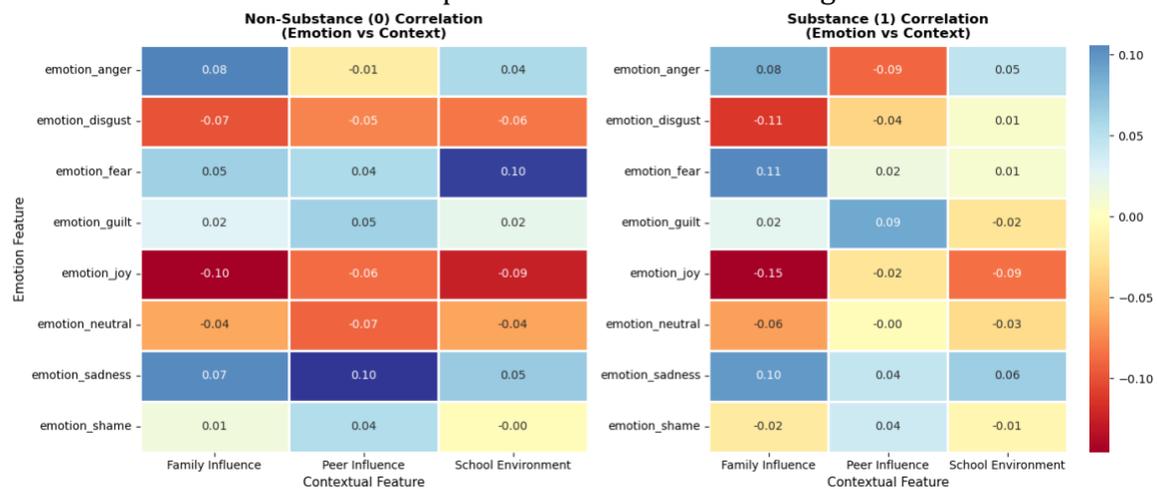

**Figure 5.** Correlations between emotional and contextual features in substance use (right) and non-substance use (left) posts.

### Predictive Modeling of Substance Use Posts

**Figure 5** shows the SHAP values for contextual and emotional features. The XGBoost model achieved an accuracy of 0.67 (recall/sensitivity = 0.75, precision = 0.68, F1 = 0.71) for substance use prediction on the test set. Higher values for joy and school environment were associated with non–substance use posts, whereas lower values indicated stronger links to substance use. In contrast, peer influence, family influence, disgust and guilt had higher importance scores for predicting substance use, underscoring their role as key emotional drivers in these discussions.

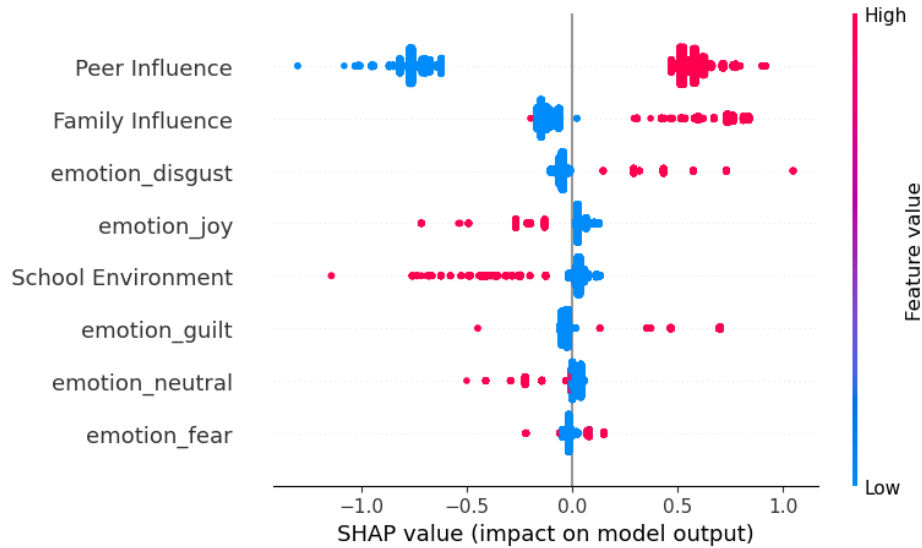

Figure 5 SHAP value showing key predictors of substance use

### Thematic Analysis of Emotion–Context Interactions

Peer-related posts revealed that fear, shame, and sadness were especially salient in the context of substance use. Teens frequently described fear of exclusion and peer pressure to conform, while shame and embarrassment emerged around public incidents (e.g., photos shared online). Joy was also prominent, linked to peer bonding during celebrations, but often intertwined with risky behaviors. These findings underscore peers as both a primary risk driver (pressure, normalization) and a key source of positive reinforcement in adolescent substance use narratives (details in Table 3 A-C).

**Table 3A.** Core Themes and Triggers for Peer-Influence Posts Across Emotions

| Emotion | Core Themes | Typical Triggers / Situations |
|---|---|---|
| **Fear** | • Social exclusion / being left out<br>• Peer pressure vs. personal values<br>• Anxiety and paranoia while using | Parties and gatherings; first-time use; fear of parental discovery; health concerns |
| **Joy** | • Peer bonding and shared experiences<br>• Celebrations and milestones<br>• Humor and light-hearted normalization | birthdays, graduation, prom; hanging out with friends; holiday parties |
| **Disgust** | • Disgust at peer pressure and normalization<br>• Unsanitary environments (bathrooms, smoke, trash)<br>• Negative perceptions of substances themselves | School culture; public intoxication; smell/taste of cigarettes and alcohol |
| **Guilt** | • Regret for actions under the influence<br>• Disappointing or deceiving family<br>• Being a "bad influence" on others | Lying to parents; boundary violations with friends; introducing younger peers to substances |
| **Shame** | • Desire to fit in with "cool" peers<br>• Embarrassing incidents while drunk/high<br>• Identity struggles and self-labeling | Peer judgment; photos/videos shared online; addiction framed as personal failure |
| **Sadness** | • Loneliness and social isolation<br>• Rejection or betrayal in relationships<br>• Mental health struggles (depression, anxiety) | Breakups; feeling excluded by friends; coping with stress through alcohol or marijuana |
| **Anger** | • Resentment of peer coercion<br>• Jealousy and competition (romantic/social)<br>• Toxic friendships and family conflicts | Friends pressuring use; rivalries over relationships; parental restrictions and double standards |

Family-related posts were dominated by fear, guilt, and sadness, reflecting the central role of parents and siblings in shaping adolescent attitudes toward substances. Posts described fear of discovery, guilt over disappointing parents, and sadness linked to conflict, divorce, or parental alcohol use. Joy was occasionally reported when families adopted permissive or celebratory attitudes toward drinking. Overall, family contexts created a dual landscape of protective expectations and risk pathways through modeling, conflict, or neglect (details in Table 3B).

**Table 3B.** Core Themes and Triggers for Family-Influence Posts Across Emotions

| Emotion | Core Themes | Typical Triggers / Situations |
|---|---|---|
| Fear | • Parental discovery & consequences<br>• Family members' substance use (modeling/normalization)<br>• Legal/health worries | Being caught by parents; school/police sanctions; living with parents/siblings who smoke/drink; fear of addiction/health harm |
| Joy | • Acceptance/ "chill" attitudes from family<br>• Bonding during sanctioned/celebratory use<br>• Exploration with family awareness | Parents allowing limited alcohol at events; holidays/birthdays; first-time trying with guidance; cultural norms around drinking |
| Disgust | • Parental hypocrisy (say "don't," but use)<br>• Negative exposure at family gatherings<br>• Aversion after bad first experiences | Family parties with heavy drinking/smoking; messy environments; witnessing harmful effects on relatives |
| Guilt | • Disappointing family/violating rules<br>• Secrecy and hiding behavior<br>• Being a bad influence on siblings/peers | Lying about use; getting caught; introducing others to substances; conflicting with family values |
| Shame | • Embarrassment when discovered by family<br>• Identity/self-image threats tied to use<br>• Regret after visible incidents | Parents find evidence; arguments at home; mental health and coping mechanisms; public fallout within family |
| Sadness | • Parental conflict/divorce & alcoholic homes<br>• Lack of support/communication<br>• Mental health struggles<br>• Desire to escape stressful family life | Chronic fighting; parental drinking; emotional neglect; grief/loss; financial stress; turning to substances for relief |
| Anger | • Trust/privacy violations & control<br>• Parental control and double standards<br>• Abuse/toxic dynamics & desire for independence | Room/phone checks; monitoring money; parents using while punishing teen use; sibling conflicts; emotional/physical abuse |

School-related posts emphasized fear, sadness, and anger as dominant emotional themes. Students expressed fear of disciplinary action (suspension, expulsion) and sadness from academic stress or social exclusion. Anger often arose from perceived unfairness, peer coercion, or frustration with dismissive school staff. Positive emotions (joy) were reported in connection with school milestones and social events, but these moments often co-occurred with substance use (details in Table 3C).

**Table 3C.** Core Themes and Triggers for School-Environment Posts

| Emotion | Core Themes (3 points) | Typical Triggers / Situations |
|---|---|---|
| Fear | • Peer pressure & exclusion risk<br>• Reputation/being judged ("uncool")<br>• Consequences (discipline/legal/health) | Parties, school searches/suspension; fear of being left out; grades stress; social media visibility |
| Joy | • Positive peer relationships/belonging<br>• Engagement in school life (sports/clubs/events)<br>• Supportive school climate & recognition | Dances, games, prom, graduation; teacher mentorship; celebrations; prosocial group norms |
| Disgust | • Normalization of use in school culture<br>• Unsanitary/toxic scenes (vaping bathrooms, intoxicated classmates)<br>• Lack of support/inaction by staff | Dirty restrooms; peers openly drunk/high; absent/ineffective prevention and response |
| Guilt | • Yielding to peer pressure against values<br>• Letting others down (parents/teachers/friends)<br>• Academic decline/self-blame | Parties and gatherings; missed assignments/exams; hiding use from adults; influencing classmates |

| Shame | • Not fitting peer norms / fear of judgment<br>• Coping with academic struggles via use<br>• Stigma/labeling of users | Rumors, photos/videos shared; poor grades; visible incidents on campus; teacher/peer scrutiny |
|---|---|---|
| Sadness | • Bullying, exclusion, loneliness<br>• Academic stress & burnout<br>• Family strain carried into school life | Cliques, rejection; exam periods; using to self-medicate depression/anxiety; family conflict spillover |
| Anger | • Peer coercion and toxic group dynamics<br>• Administrative inaction/unfair rules<br>• Stigma/judgment blocking help-seeking | Friends pushing use; perceived double standards; scarce counseling/coping education; punitive policies |

## Discussion

This study examined emotional and contextual correlates of adolescent substance use in social media posts using statistical tests, large language models, and latent thematic analysis. Three broad conclusions emerge. First, substance-related discussions displayed clear seasonal variation, peaking during school breaks and holidays, consistent with prior research on youth mental and behavioral health cycles [47–49]. Such peaks may reflect heightened social activity and stressors (e.g., loneliness, family tension) that increase vulnerability to substance user. Second, sadness was the most prevalent emotion in substance use posts, aligning with evidence that negative affect predicts adolescent substance involvement [50,51]. Third, peer influence was the dominant contextual driver, followed by family and school factors; correlation patterns indicated that guilt, sadness, and fear clustered strongly with peer contexts, consistent with adolescent socio-emotional reactivity and peer susceptibility during puberty [52,53].

Negative emotions played distinct roles. Guilt, reflecting concern for the impact of one's behavior on others, often co-occurred with remorse and regret [56]. In our data, guilt was linked to corrective intentions following use, suggesting a potentially protective function. By contrast, shame was more closely tied to threats to self-worth and concealment, associated with risky dynamics such as peer validation and identity performance[55]. This distinction aligns with prior research indicating that guilt may foster reparative action, whereas shame amplifies vulnerability through social identity pressures [56,57]. Disgust frequently emerged as internal conflict or peer critique, sometimes serving as a moral boundary, but it weakened in contexts where use was normalized.

Contextual patterns further highlighted the central role of peers, consistent with extensive evidence that peer norms strongly shape adolescent substance involvement [58]. Prior work emphasizes the need to distinguish peer selection from peer socialization when developing prevention strategies [59]. In our data, peer influence was most strongly associated with guilt in substance-use posts, suggesting that adolescents often frame their use through interpersonal responsibility and regret. Family influence was uniquely tied to fear and sadness, highlighting concerns about parental discovery and strained relationships. School contexts, in contrast, were more closely linked to sadness in substance-use posts but to fear in non–substance-use posts, pointing to different stress pathways across environments. These findings extend earlier work showing that adolescents' heightened emotional reactivity and reduced regulatory capacity amplify peer-context risks [60].

LLM-assisted thematic coding provided further nuance. Sadness was tied to loneliness and social disconnection—even "in a crowd"—underscoring belonging as a core motive. Fear functioned ambivalently: fear of exclusion promoted conformity, whereas fear of parental sanction or long-term harm deterred use. Disgust served as a moral boundary (e.g., disdain for vaping peers), but its protective effect weakened in cultures where use was normalized. Guilt reflected the intention–behavior gap, where reflective goals were overridden by situational impulses, prompting self-reproach. Shame often followed use but also fueled a feedback loop of risky behavior through identity performance ("being cool") and later regret [61].

Family and school environments were similarly pivotal. Family-related risks included parental conflict, divorce, substance use, and low support, while strong bonds and positive adult models emerged as protective[62]. In schools, triggers included academic stress, parties, and safety concerns, while policies and supportive programs provided buffers. Addressing academic pressures through stress-management strategies may reduce risk.

Together, these findings advance understanding of adolescent substance use by identify peers as both primary risk drivers and potential sources of support, families as contexts of both conflict and protection, and schools as key environments shaping stress and coping. At the individual level, the contrast between disgust and guilt underscores how discrete emotions channel risk versus resilience pathways[56]. These insights extend theory on the socio-emotional mechanisms of adolescent substance use while pointing to multi-level targets for prevention.

## Limitations

This study has several limitations. First, the anonymous nature of Reddit prevents control over demographic variables such as age, gender, ethnicity, or location, which constrains the generalizability of the findings. Moreover, it is not possible to verify that all posts were authored by teenagers; contributions from adults or bots may have introduced bias. Second, the use of GPT-4 for emotional and contextual annotations introduces reliance on large language model parameters, which are subject to inherent limitations. Although manual validation was performed, annotation accuracy ultimately depends on prompt design and model capabilities. Third, the comparative design, which involved randomly sampling non–substance-use posts, may not fully capture broader discourse patterns, and the absence of longitudinal data restricts insight into the developmental course of substance use behaviors. Finally, the self-reported nature of the data may involve exaggeration or omission of substance use, further limiting interpretability. Future research should incorporate longitudinal designs and larger, more diverse datasets to strengthen representativeness and analytic robustness.

## Conclusions

In conclusion, our study highlights disgust and guilt as key emotional correlates of adolescent substance use the differentiated roles of guilt and shame, the ambivalent function of fear, and the salience of sadness as core emotional correlates of adolescent substance use. Peers emerged as the dominant contextual driver, with family and school also important while family and school contexts contributed distinct risk and protective dynamics. Future research should further validate LLM-based annotation methods and expand to other platforms~ refine and validate LLM-based annotation for emotion–context mechanisms and extend analyses to other youth-oriented platforms. However, together, these findings suggest that multi-level prevention programs addressing peer, family, school, and individual factors are needed advance theory on the socio-emotional mechanisms of adolescent substance use and point to multi-level leverage points for future prevention.

## Conflicts of Interest
None declared.

## Data Availability
The data used in this study are publicly available for download at the following link: [RedditData](#).

## References

[1] National Center for Drug Abuse Statistics. (2023). Drug use among youth: Facts & statistics. Drug Abuse Statistics. Retrieved November 12, 2024, from https://drugabusestatistics.org/teen-drug-use/